\documentclass{article}

    \usepackage[final,nonatbib]{neurips_2020}

\usepackage[utf8]{inputenc} 
\usepackage[T1]{fontenc}    
\usepackage{hyperref}       
\usepackage{url}            
\usepackage{booktabs}       
\usepackage{amsfonts}       
\usepackage{nicefrac}       
\usepackage{microtype}      

\usepackage{amsmath}
\usepackage{amssymb}

\usepackage{graphicx}
\usepackage{subcaption}
\usepackage{textcomp}
\usepackage[per-mode=symbol-or-fraction]{siunitx}

\usepackage{booktabs}

\usepackage[capitalise]{cleveref}

\usepackage{svg}
\usepackage[square,numbers]{natbib}

\title{Optimising Placement of Pollution Sensors  \\in Windy Environments}

\author{%
  Sigrid Passano Hellan \quad Christopher G. Lucas \quad Nigel H. Goddard \\ 
  School of Informatics\\
  University of Edinburgh, UK\\
  \texttt{\{s.p.hellan, c.lucas, nigel.goddard\}@ed.ac.uk} \\
}

\begin{document}

\maketitle

\begin{abstract}

Air pollution is one of the most important causes of mortality in the world. Monitoring air pollution is useful to learn more about the link between health and pollutants, and to identify areas for intervention. Such monitoring is expensive, so it is important to place sensors as efficiently as possible. Bayesian optimisation has proven useful in choosing sensor locations, but typically relies on kernel functions that neglect the statistical structure of air pollution, such as the tendency of pollution to propagate in the prevailing wind direction. We describe two new wind-informed kernels and investigate their advantage for the task of actively learning locations of maximum pollution using Bayesian optimisation.

\end{abstract}

\newcommand{\pmt}{PM\textsubscript{2.5} }
\newcommand{\NOt}{NO\textsubscript{2} }

\newcommand{\pmtns}{PM\textsubscript{2.5}}
\newcommand{\NOtns}{NO\textsubscript{2}}

Air pollution is one of the most important causes of death globally. 
Particulate matter with diameter less than \SI{2.5}{\micro\meter}, \pmtns, 
is estimated to cause 3-4 million deaths yearly \cite{cohen_estimates_2017,lelieveld_contribution_2015}. 
Through monitoring one can learn more about the impact of air pollution on public health and identify areas requiring intervention.
Traditional monitoring networks consist of a few high-quality reference sensors \cite{carminati_emerging_2017}, but many low-cost pollution sensors are now available, e.g. \cite{liu_low-cost_2020},
which can be used to build higher-resolution 
models to complement the traditional networks.   
This work aims to improve monitoring networks by improving the placement of low-cost, potentially noisy sensors to monitor average concentrations of \pmt and identify areas of high average concentration, assuming that sensors can be placed incrementally.
This problem -- of finding the maxima of a black-box function given sparse and noisy observations -- is well-matched to Bayesian optimisation (BO).
Most machine learning research on air pollution tends not to consider the problem of efficient sensor placement, e.g. \cite{smith_machine_2019,liu_data_2016,liu_modeling_2018,fei_accurate_2020,wen_novel_2019}. 
To our knowledge, previous work on environmental sensing using BO and related methods, e.g. \cite{marchant_bayesian_2012,morere_sequential_2017,singh_modeling_2010}, has not considered wind.

Wind is known to move pollutants downwind \citep[p.~78-80]{watson_atmospheric_1988}, and work on gas mapping has shown the importance of incorporating this effect \cite{reggente_using_2009,asenov_active_2019}. For instance,
\cite{asenov_active_2019} used a gas localisation task to show that  a method incorporating wind velocity outperforms BO using wind-insensitive kernels. Because we are focusing on time-averaged pollution concentrations, detailed wind information and fluid simulations are not essential: The covariance structure of pollution is driven by the prevailing wind direction, which can be characterised with a single parameter.
The kernels presented here allow such information to be incorporated into BO, and have the advantage that wind need not be measured separately. This is desirable from a cost perspective, as additional wind sensors are not needed.

The key question is whether these kernels improve our ability to find the location(s) of highest pollution in an area, with a minimum number of sensor placements, i.e., samples of time-averaged pollution concentrations. To answer this question using real data, we used satellite images of nitrogen dioxide (\NOtns) concentrations (see Supplementary Material for details). 
\NOt is a health-related pollutant in its own right \cite{luo_acute_2016}, and we assume that the dispersion of \pmt is comparable. 
While satellite data cannot replace surface-level sensor data, it provides a good testbed given the spatial regularity and density of the observations, and it allows us to set aside complex effects of urban topography.

\section{Wind-informed kernels for Bayesian optimisation}
\newcommand{\ucb}{\alpha{_\mathrm{UCB}}}
\newcommand{\data}{\mathcal{D}}

In Bayesian optimisation \cite{shahriari_taking_2016} a probabilistic model, in our case a GP, is fitted to the set of observed data points. This model is then used to compute an acquisition function, and the next sample is taken at the point which maximises this function. The new sample is added to the set of data points and the process is repeated. 
In this paper, the upper confidence bound (UCB) \cite{srinivas_gaussian_2012} was used as the acquisition function,
$\ucb(\boldsymbol{x};\beta,\data,M)~=~\mu(\boldsymbol{x};\data,M)~+~\beta \sigma(\boldsymbol{x};\data,M)$, with $\beta=1$. $\mu(\boldsymbol{x};\data,M)$ and $\sigma(\boldsymbol{x};\data,M)$ are the posterior mean and standard deviation of a point $\boldsymbol{x}$ given data $\data$ and model $M$. 
An entropy-based acquisition function \cite{hernandez-lobato_predictive_2014} and other values of $\beta$ were tested, without marked improvement. 
To compute $\mu(\cdot)$ and $\sigma(\cdot)$, covariances need to be computed between the data points using a kernel function. 
A thorough introduction to GPs can be found in \cite{rasmussen_gaussian_2006}.

The motivation for the wind-informed kernels is to exploit the directional structure of wind-driven pollution transport. 
Intuitively, one would expect to see the highest levels of pollution downwind from a source, and if one recorded a high pollution concentration in a location one would expect to find a source upwind from it.
In this work we do not aim to locate sources, but to look at concentrations, and so do not differentiate between upwind and downwind movement.

\newcommand{\sigmaR}{\sigma_R}
\newcommand{\sigmaS}{\sigma_D}
\newcommand{\sigmaP}{\sigma_P}
\newcommand{\ldir}{l_D}

\newcommand{\tb}{\boldsymbol{\tau}}
\newcommand{\covaria}{k_S \left(\tb;l,\ldir,\sigmaR,\sigmaS,\gamma \right)}
\newcommand{\covariashort}{k_S \! \left(\tb;\theta_S \right)}
\newcommand{\covprod}{k_P \left(\tb;l,\ldir,\sigmaP,\gamma \right)}
\newcommand{\covprodshort}{k_P  \!\left(\tb;\theta_P \right)}
\newcommand{\rbf}{k_R \left(\boldsymbol{\tau};l,\sigmaR \right)}
\newcommand{\rbflow}{k_R (\boldsymbol{\tau};l,\sigmaR )}
\newcommand{\rbfhat}{k_R \left(\boldsymbol{\tau};l,\sigmaP \right)}
\newcommand{\eye}{\boldsymbol{\mathbb{I}}}

\begin{figure}[ht]
  \centering
  \includegraphics[width=0.65\linewidth]{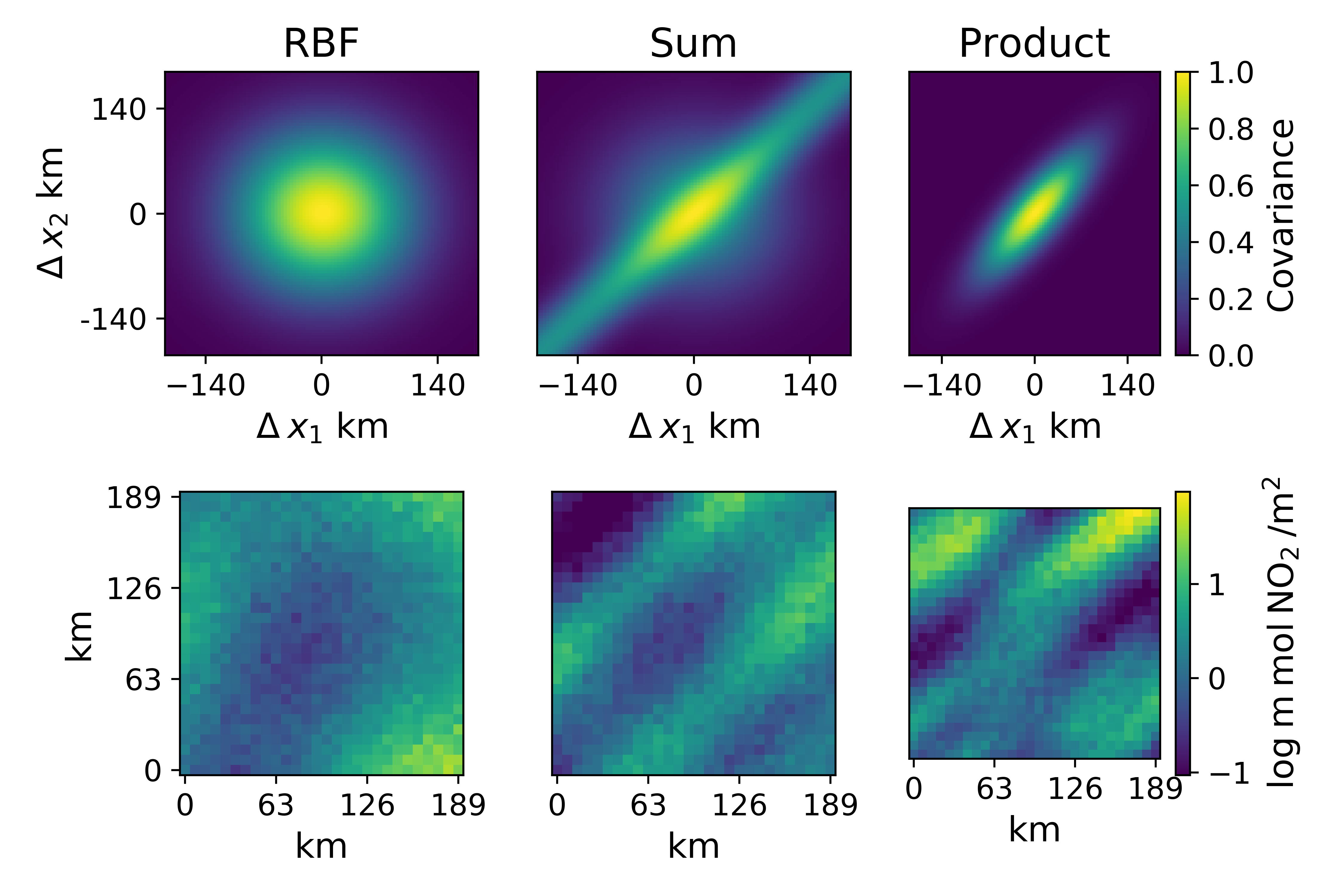}
  \caption{Visualisation of wind-informed kernels from \cref{eq-novel-sum,eq-novel-product}. The top row shows covariance as function of distance, and the bottom row shows example images sampled from the kernels.  $\sigmaS^2=0.5$, $l=105$, $\ldir=35$,
  $\gamma=\pi/4$, $\sigmaP^2=1$, $\sigma_n^2=0.01$. For Sum $\sigmaR^2=0.5$, for RBF $\sigmaR^2=1$.}
  \label{fig:kernel-visualisation}
\end{figure}

The standard radial basis function (RBF) was modified to create the wind-informed kernels.
The RBF kernel calculates the covariance between two points $\boldsymbol{x}_i$ and $\boldsymbol{x}_j$ as $\rbflow~\!=\!~\sigmaR^2 \exp ( -\tb^T \tb/l^2)$ where $\tb=\boldsymbol{x}_i - \boldsymbol{x}_j$, $l$ is the length scale and $\sigmaR^2$ the signal variance.
The new kernels $k_S(\cdot)$, \emph{Sum}, and $k_P(\cdot)$, \emph{Product}, are given in \cref{eq-novel-sum,eq-novel-product} and visualised in \cref{fig:kernel-visualisation}. The kernels are a combination of a standard RBF kernel and another RBF kernel on the distance orthogonal to the wind direction. This distance is given by $\tilde{\tau} \left(\tb;\gamma \right) = ||\tb - \left(\tb^T \boldsymbol{b} \right)\boldsymbol{b} ||$ where $\boldsymbol{b} = [\cos(\gamma) \:\: \sin(\gamma)]^T$ and $\gamma$ is the wind direction hyperparameter. 
Further,
$\ldir$ is the directional length scale, $\sigmaS^2$ the directional signal variance, $\sigmaP^2$ the signal variance for $k_P(\cdot)$, 
$A~=~\begin{bmatrix}{} \sin(\gamma)^2 &  - \sin(\gamma)\cos(\gamma)  \\ - \sin(\gamma)\cos(\gamma) & \cos(\gamma)^2 \end{bmatrix} $ and 
$ E= \ldir^2/l^2 \eye + A $. 
Finally, $\theta_S =\{l, \ldir, \sigmaR, \sigmaS, \gamma \}$ and $\theta_P =\{l, \ldir, \sigmaP, \gamma \}$ denote sets of hyperparameters.
The kernel hyperparameters, together with the noise hyperparameter $\sigma_n^2$, are fitted in the GP tuning step. 

\begin{align}
    \covariashort \!=& \rbf \!+ \! k_R\! \left(\tilde{\tau}  \left(\tb;\gamma \right);\ldir, \sigmaS  \!\right) 
    \!= \! \sigmaR^2 \exp  \!\left( \frac{- \tb^T \tb}{l^2}   \!\right)\!+ \!\sigmaS^2 \exp  \!\left( \frac{-\tb^T A \tb}{\ldir^2}  \!\right) \label{eq-novel-sum} \\
    \covprodshort \! =& \rbfhat k_R \left(\tilde{\tau}  \left(\tb;\gamma \right);\ldir, 1 \right) = \sigmaP^2 \exp \left( \frac{-\tb^T E \tb}{\ldir^2} \right) \label{eq-novel-product} 
\end{align}

These kernels are sums and products of RBF kernel functions that are valid for all inputs, and are thus valid kernel functions themselves
\citep[p.~83,95]{rasmussen_gaussian_2006}.

\section{Data}

\begin{figure}[ht]
\begin{subfigure}[b]{.32\textwidth}
  \centering
  \includegraphics[width=0.95\linewidth]{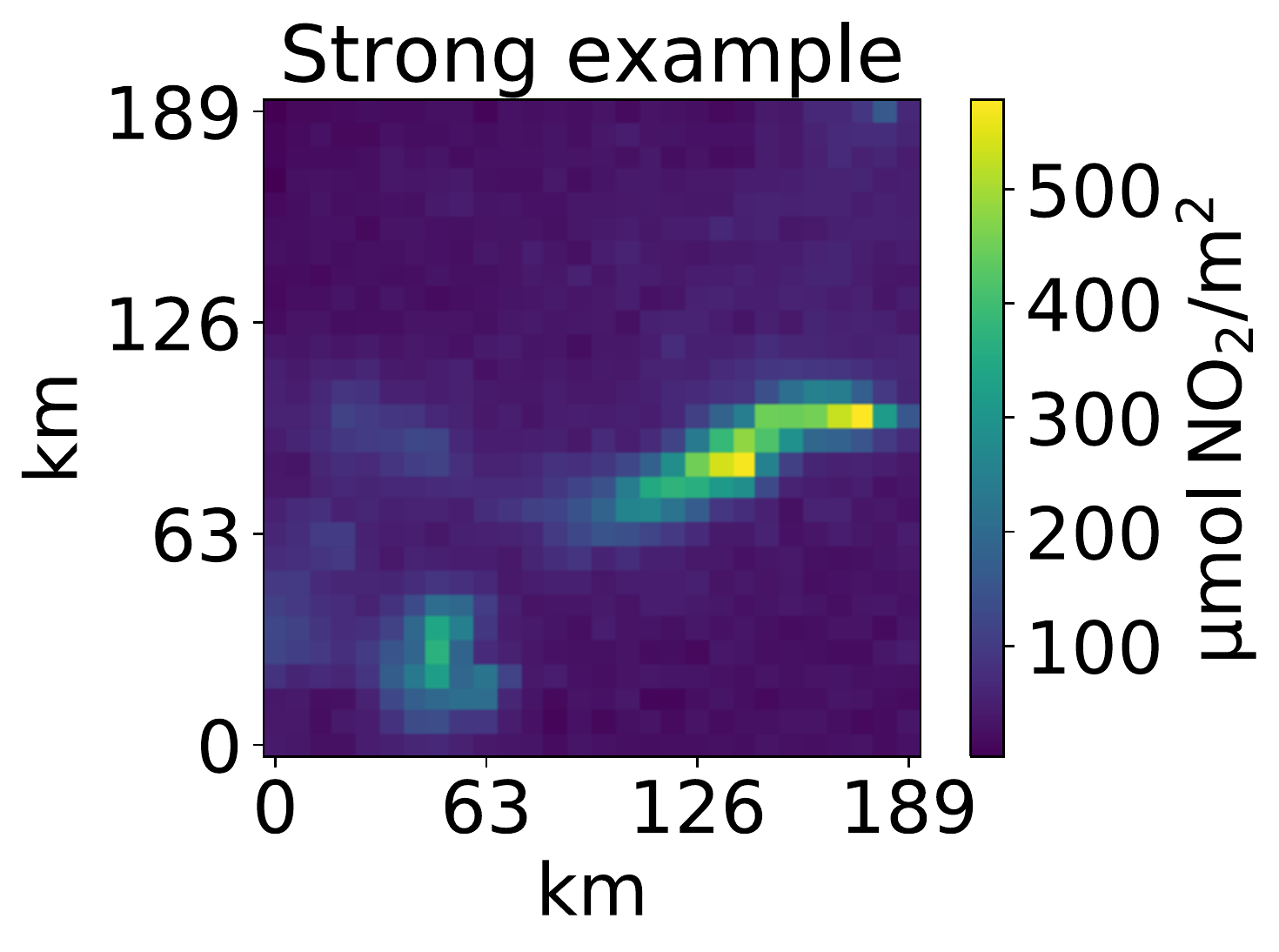}
  \label{fig:problem-example-strong}
\end{subfigure}%
\begin{subfigure}[b]{.32\textwidth}
  \centering
  \includegraphics[width=0.95\linewidth]{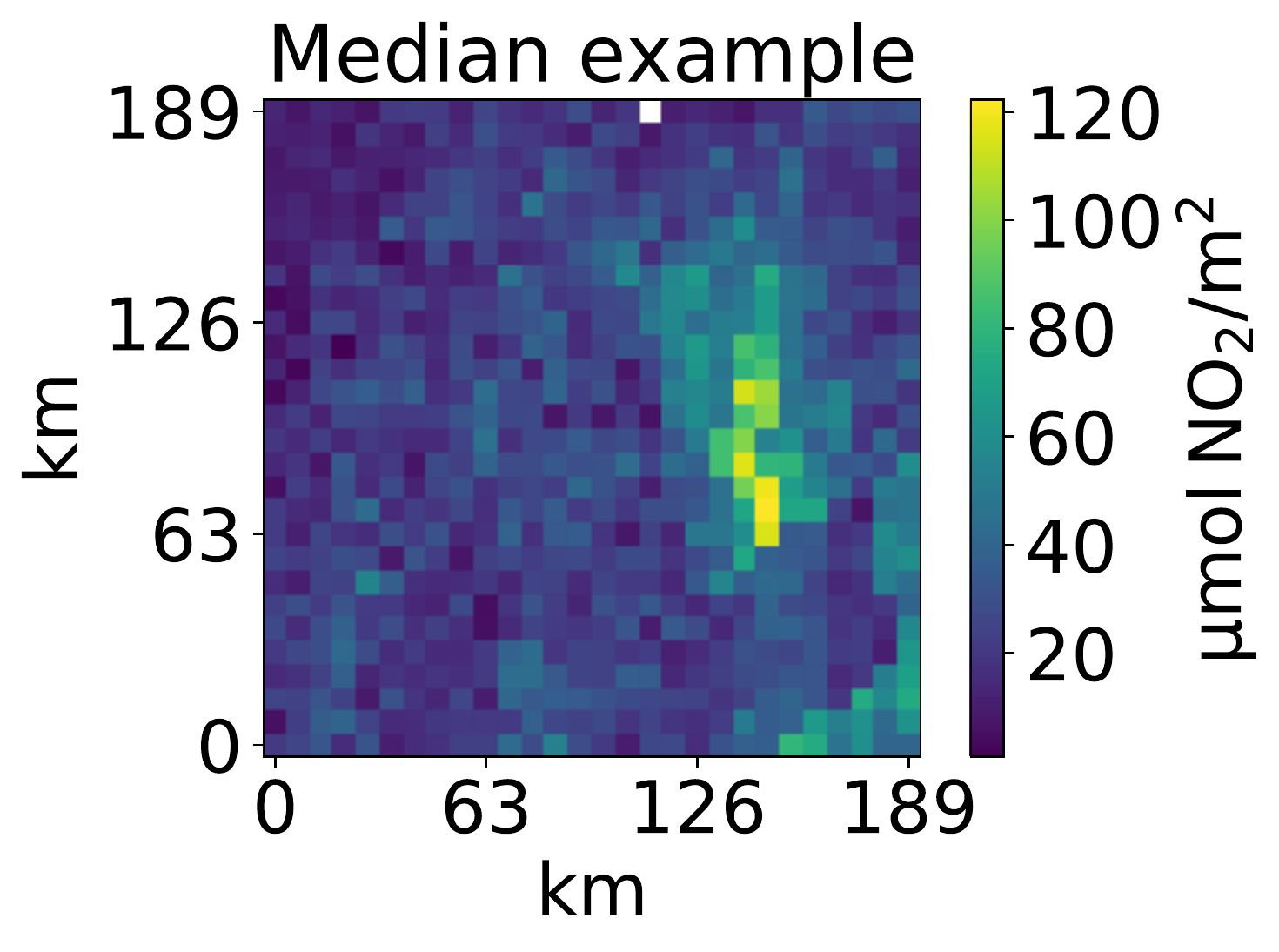}
  \label{fig:problem-example-median}
\end{subfigure}%
\begin{subfigure}[b]{.32\textwidth}
  \centering
  \includegraphics[width=0.92\linewidth]{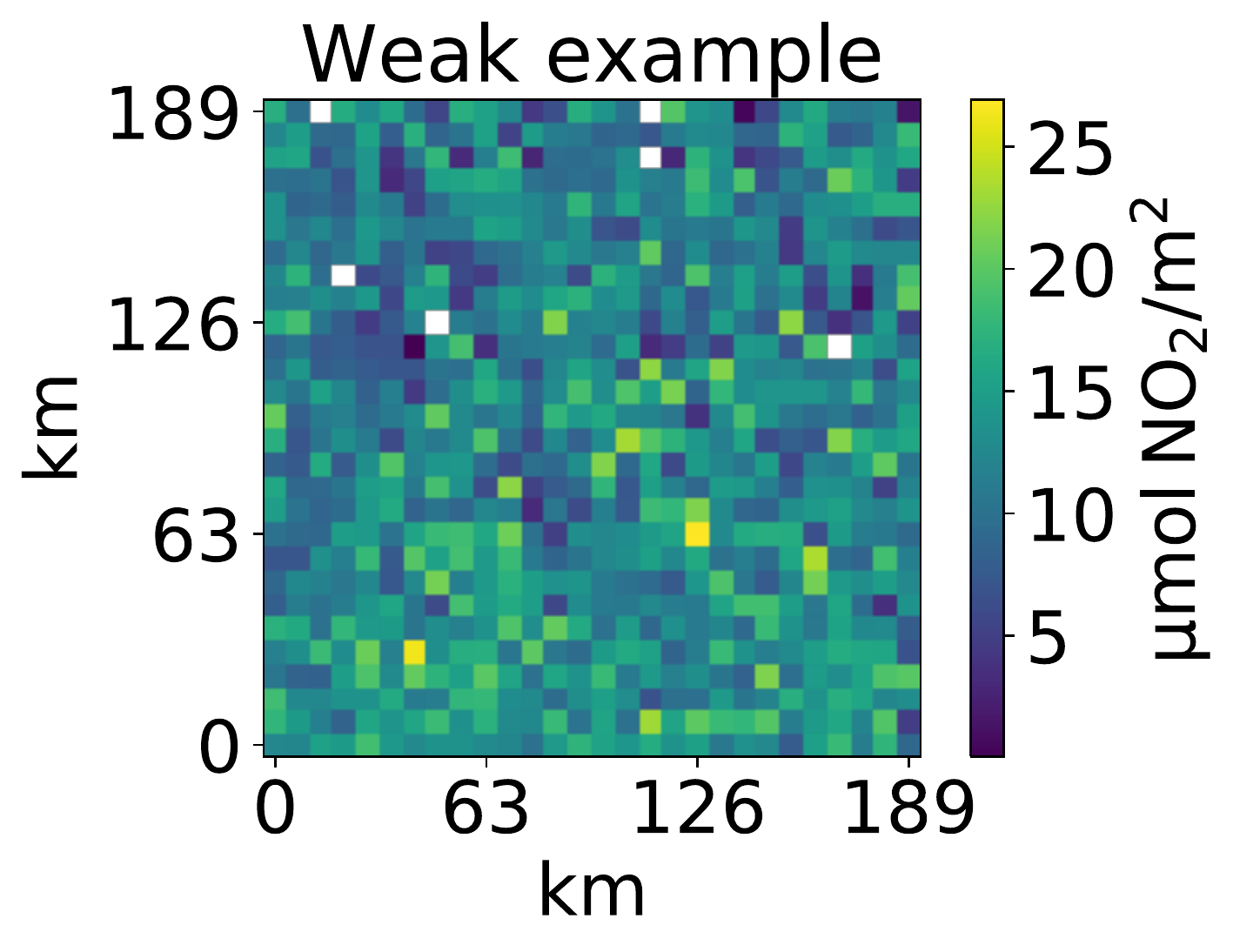}
  \label{fig:problem-example-weak}
\end{subfigure}%
\caption{Example \NOt concentration images from the Strong, Median and Weak subsets. White pixels indicate missing data. Note the difference in scale between the subsets, and the pollution plumes from point sources in two of the examples. Such plumes were present in many images, particularly in the Strong subset, and are what the wind-informed kernels are meant to capture.}
\label{fig:problem-example}
\end{figure}

The data set consists of 1083 images of 28x28 pixels, with each pixel giving the \NOt concentration in mol/m\textsuperscript{2} within a ca. 7x7km box on the ground reaching 
the upper troposphere. 
The images are from October and November 2018 and cover landmasses across the world. 
More information is given in the 
Supplementary Material.
We removed images with more than 10 \% missing values, and split the data into subsets. 
The split was done as \NOt concentrations varied greatly between images,
and locating peaks of high pollution is more important. 
Three subsets were created as follows: the images were ordered by the highest \NOt concentration present, and the top, median and bottom 50 images put into sets called 'Strong', 'Median' and 'Weak', respectively. See \cref{fig:problem-example}.
Three adjacent sets of 10 images each were created for tuning and normalisation. A final set of 100 images, dubbed 'Selection', 
was created by picking every ninth of 
the remaining ordered images. 
The first three subsets were created to inspect performance on different degrees of difficulty, and the last to test the overall performance.

\section{Results} \label{sec-results}


We used log concentrations rather than raw concentrations as our data -- and pollution concentrations in general \cite{cats_prediction_1980} -- are approximately log-normal, satisfying the assumption of Gaussian marginal and residual distributions in GP models. Images were normalised using means and standard deviations from the associated tuning set. The rationale for this was that the true mean and standard deviation for a deployment region would likely be unknown, but values from similar areas available. 
The tuning sets were also used to develop  priors for the hyperparameters, except for $\gamma$ for which a uniform prior was used. 
For each tuning image, gradient descent with 200 hyperparameter initialisations was used to maximise the log marginal likelihood, $\log(p(Y^i|M))$, where $M$ denotes the model, including kernel and hyperparameter choices, and $Y^i$ is the pixel values in image $i$. 
A log normal distribution was then fitted to these hyperparameter samples using the maximum likelihood estimator and this was used as a prior. 
For the Selection subset, the Median tuning set was used for normalisation and prior computation.
The models obtained on the tuning sets were compared using the Bayesian information criterion (BIC) \cite{schwarz_estimating_1978}. It was highest for the Sum kernel, followed by the Product kernel, c.f. the Supplementary Material. 
This shows that the wind-informed kernels are able to capture more of the structure of the pollution in the images. For the Weak subset little difference was seen.

\begin{figure}
\centering
\begin{subfigure}[b]{.47\textwidth}
  \centering
  \includegraphics[width=.95\linewidth]{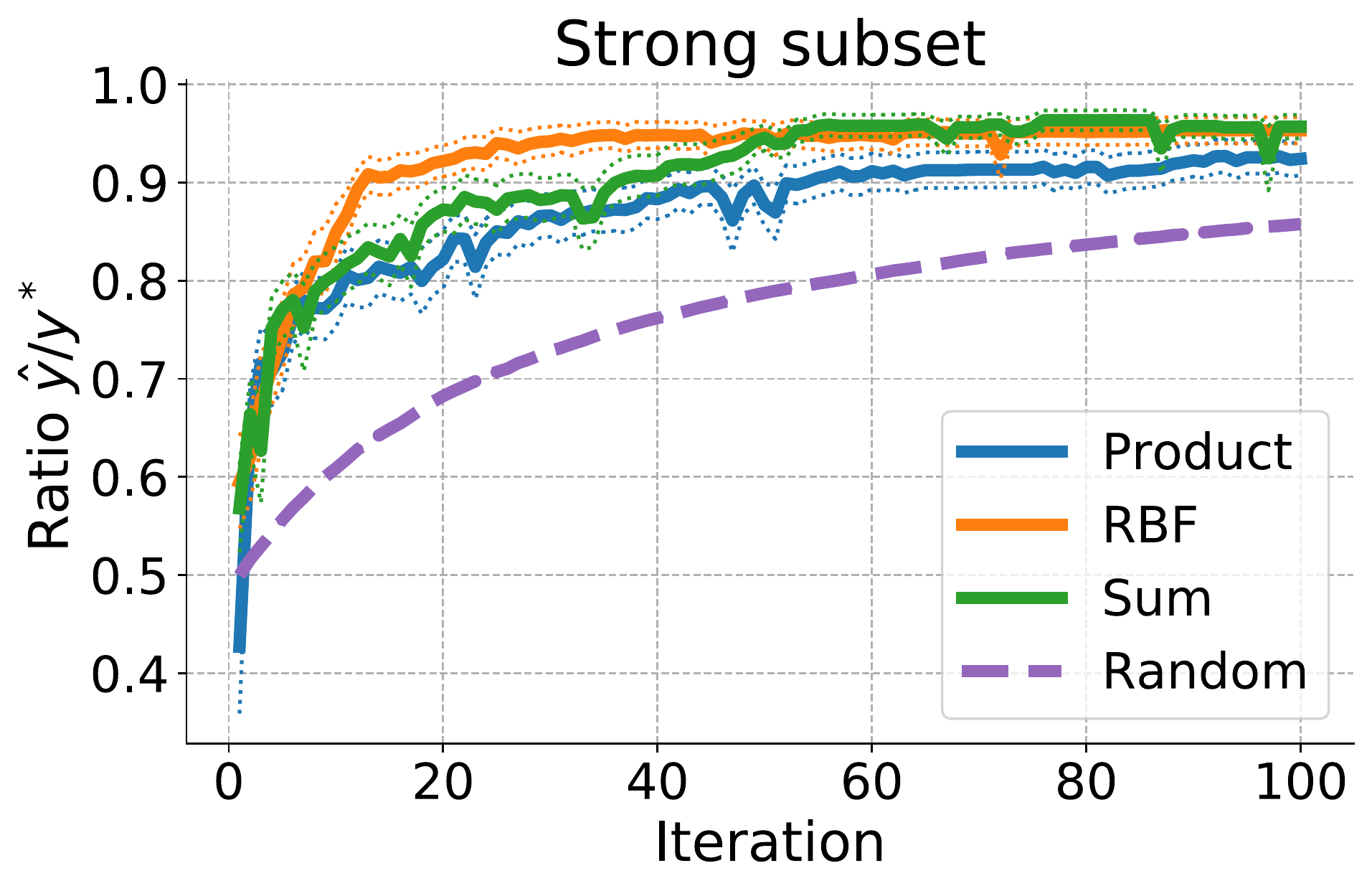}
  \label{fig:results-strong-true}
\end{subfigure}%
\begin{subfigure}[b]{.47\textwidth}
  \centering
  \includegraphics[width=.95\linewidth]{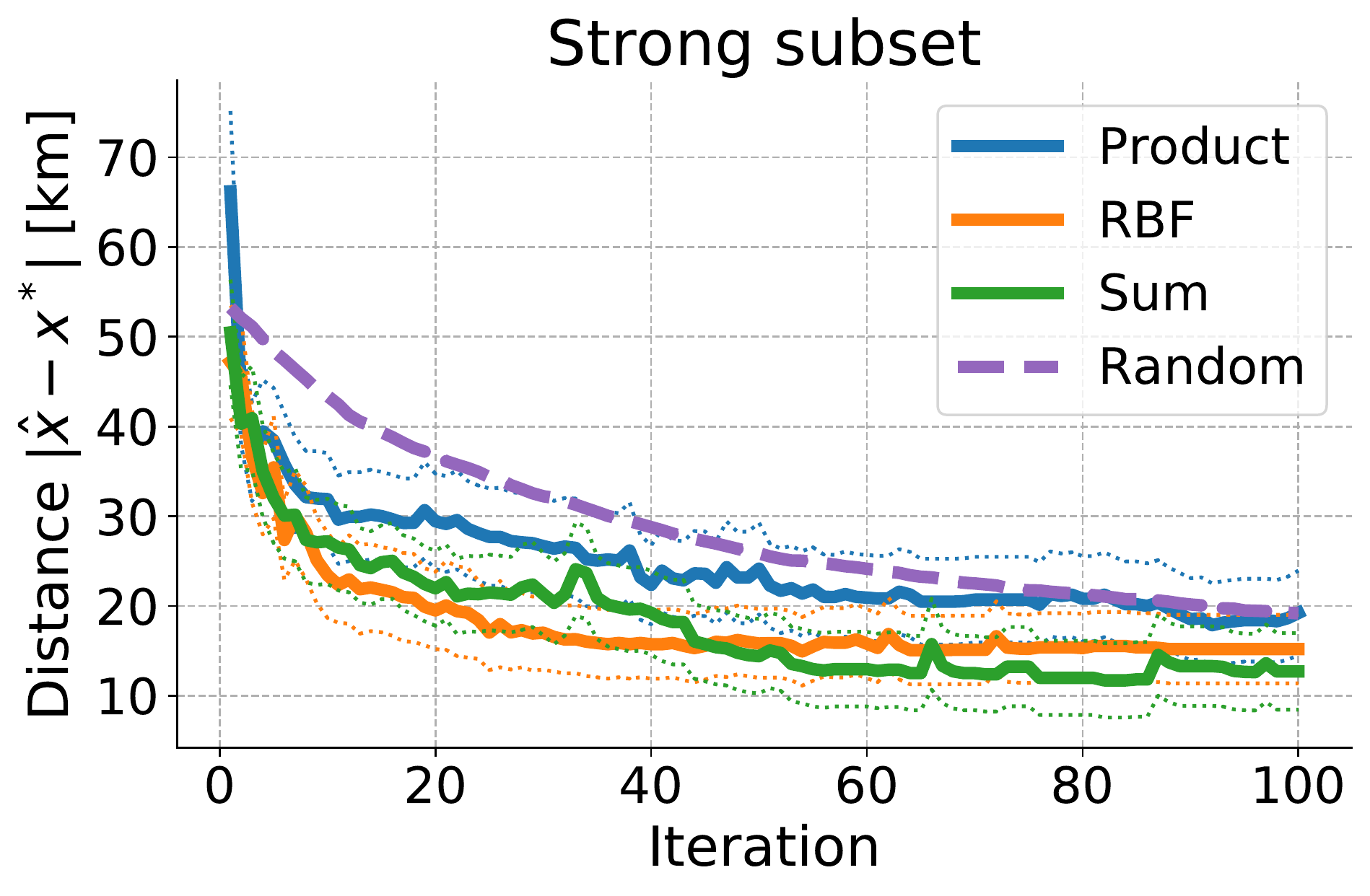}
  \label{fig:results-strong-dist}
\end{subfigure}
\begin{subfigure}[b]{.47\textwidth}
  \centering
  \includegraphics[width=.95\linewidth]{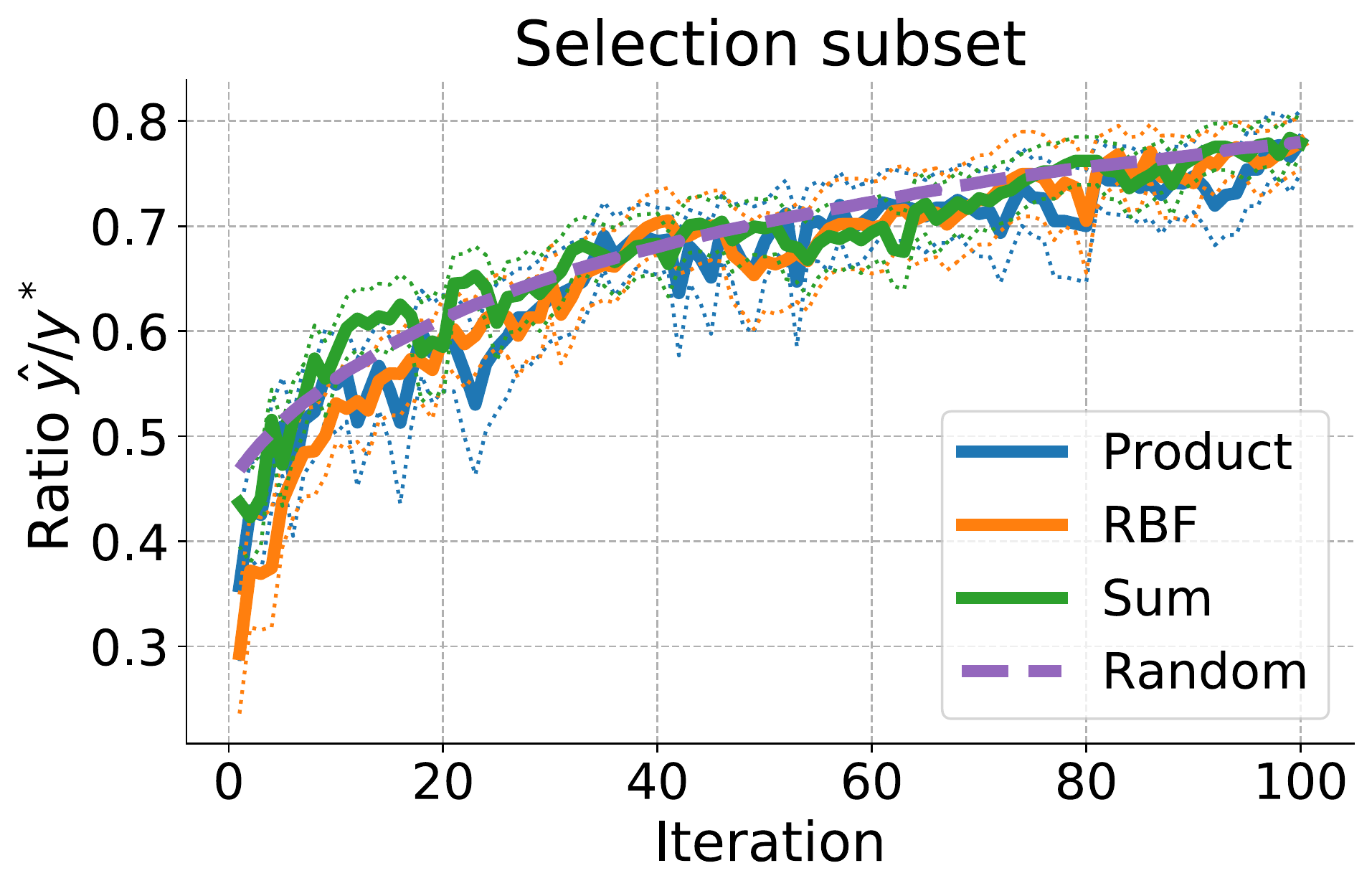}
  \label{fig:results-selection-true}
\end{subfigure}%
\begin{subfigure}[b]{.47\textwidth}
  \centering
  \includegraphics[width=.95\linewidth]{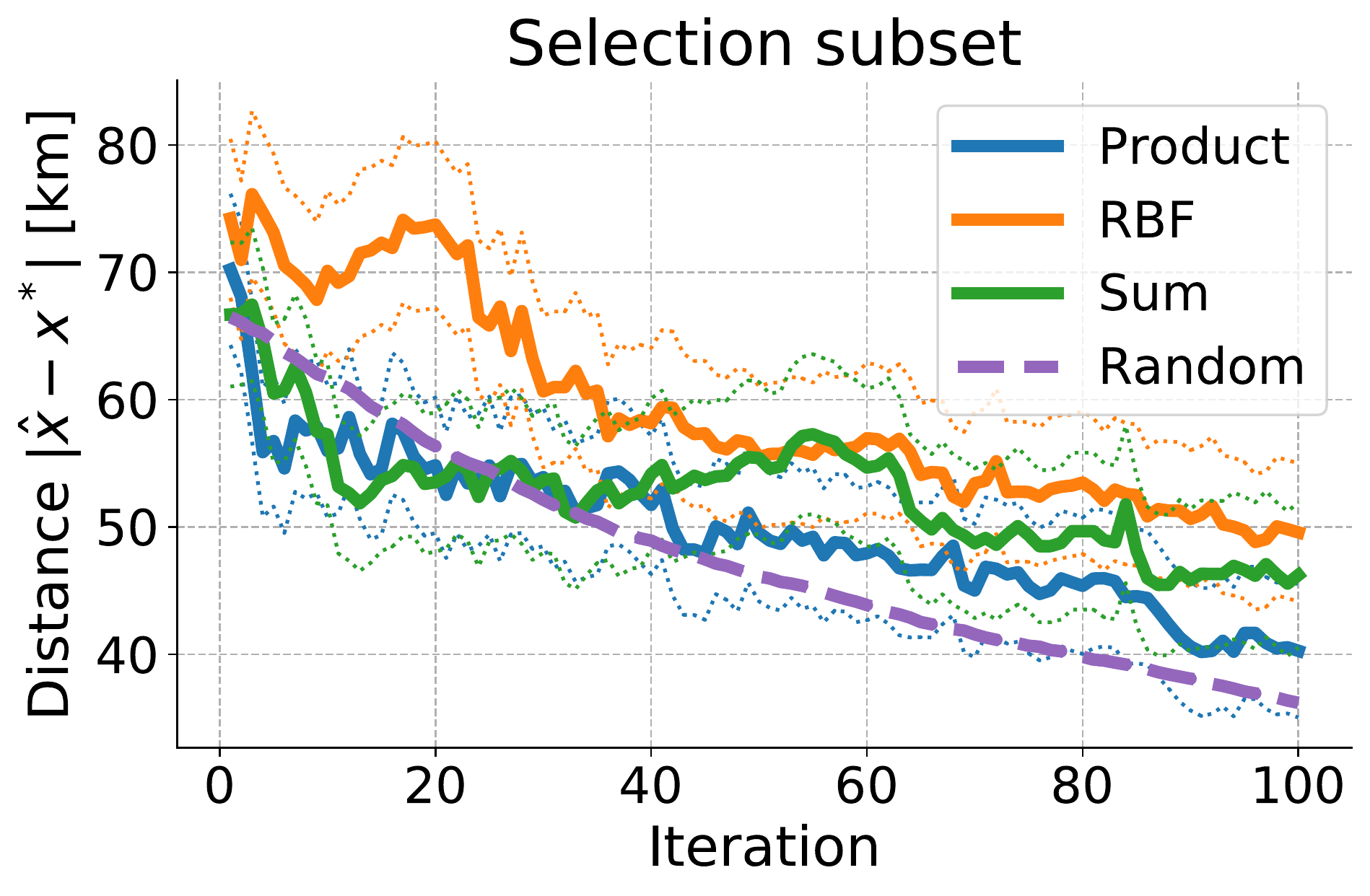}
  \label{fig:results-selection-dist}
\end{subfigure}%
\caption{Comparison of kernel performance on Strong and Selection subsets. $\hat{x}$ is the estimated maximiser and $x^*$ the true maximiser. $\hat{y}$ and $y^*$ are the true concentration values at $\hat{x}$ and $x^*$, respectively. Both metrics are shown as means with standard deviation of the mean in dotted lines. 
The Sum kernel reaches slightly better values than the RBF kernel on the Strong subset, and the Product kernel outperforms the other kernels on the distance metric on the Selection subset.}
\label{fig:results-strong}
\end{figure}

\begin{figure}
  \centering
  \includegraphics[width=0.92\linewidth]{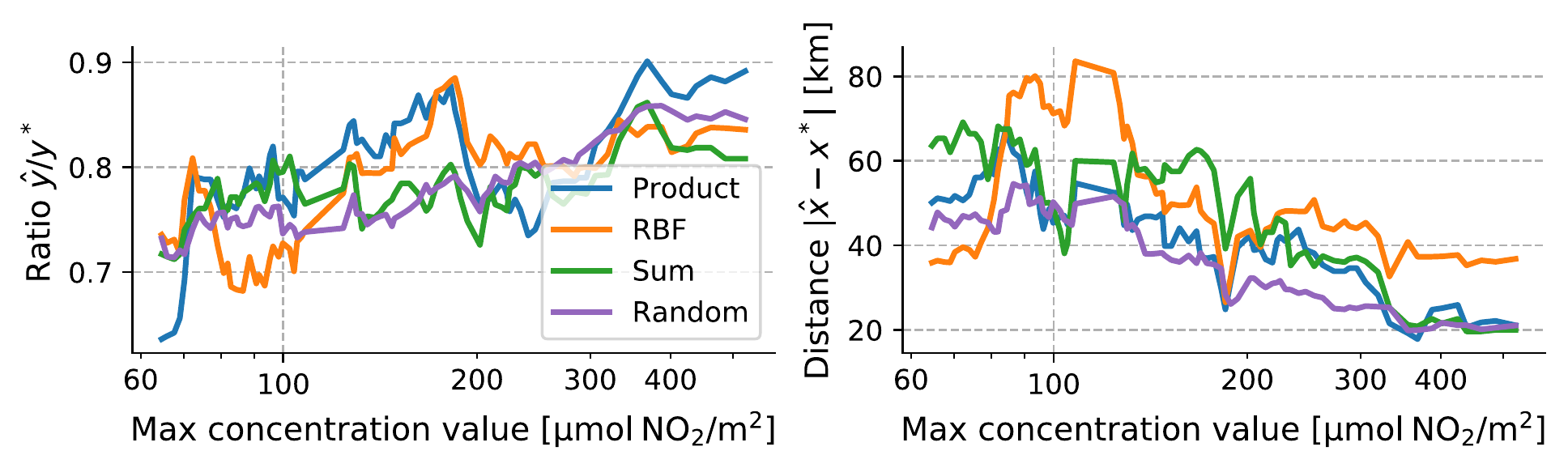}
  \caption{Inspection of performance after 100 iterations over the range of concentrations in the Selection subset. Each line shows the running average of the 20 previous values. $\hat{x}$, ${x^*}$, $\hat{y}$ and $y^*$ as defined for \cref{fig:results-strong}. As can be seen the Product kernel outperforms RBF across most concentration.}
  \label{fig:results-range}
\end{figure}

The kernels were evaluated on the task of locating the pixel with the maximum pollution concentration. Results on two of the subsets, Strong and Selection, are presented.  For comparison, our baselines were (1) Bayesian optimisation with an RBF kernel (RBF) and (2) a policy of choosing points uniformly at random (Random). The latter was repeated 100 times per image and the mean taken. The experiments were initiated with 10 randomly selected pixel values.
\cref{fig:results-strong} shows the performance as a function of iterations done,
showing the Sum kernel performs better than the RBF kernel on the Strong subset.
For the selection subset, the Random baseline beats the RBF baseline. This indicates that the problem is hard for BO, as would be expected from the example images in \cref{fig:problem-example}. The Product kernel outperforms the other kernels on the distance metric, while the ratio metric is very similar for all methods. 
\cref{fig:results-range} looks more closely at the performance on the Selection subset as a function of the maximum pollution concentration present in each image. 
It shows that the performance improves with higher levels of pollution and also that over most of the concentration values the Product kernel is better than the RBF.

\section{Discussion and future work}

By augmenting a standard RBF kernel with a directional element that captures the prevailing wind direction, we can better capture the covariance structure in \NOt pollution, as indicated by the BIC scores. The results also show that this better estimates pollution maxima and the locations of those maxima for some of the images. 
The improvement was less clear than we expected from the BIC scores. 
We hypothesise that this is due to the hyperparameter tuning,
which the results were sensitive to. For an example, see the 
Supplementary Material.
Therefore, we plan to improve our priors by basing them on a larger tuning set and using hierarchical modelling.
Additionally, the wind-informed kernels should be tested on more data sets, and particularly on data sets of urban air pollution. As wind is also important in moving air pollution over short distances one would expect improvement, but this might be distorted by other factors, like buildings.
The approach used, of incorporating a data transformation into the covariance function, can be extended to other kernels, e.g. the set of Mat\'{e}rn kernels, or combined in other ways, e.g. a combination of Sum and Product kernels.

\begin{ack}
The authors would like to thank Dr Douglas Finch at the University of Edinburgh for extracting the data and helpful correspondence. 

This work was supported in part by the EPSRC Centre for Doctoral Training in Data Science, funded by the UK Engineering and Physical Sciences Research Council (grant EP/L016427/1) and the University of Edinburgh.
\end{ack}

\section*{References}

\medskip

{\small
\bibliography{main}

\begin{thebibliography}{22}
\providecommand{\natexlab}[1]{#1}
\providecommand{\url}[1]{\texttt{#1}}
\expandafter\ifx\csname urlstyle\endcsname\relax
  \providecommand{\doi}[1]{doi: #1}\else
  \providecommand{\doi}{doi: \begingroup \urlstyle{rm}\Url}\fi

\bibitem[Cohen et~al.(2017)Cohen, Brauer, Burnett, Anderson, Frostad, Estep,
  Balakrishnan, Brunekreef, Dandona, Dandona, Feigin, Freedman, Hubbell,
  Jobling, Kan, Knibbs, Liu, Randall, and Forouzanfar]{cohen_estimates_2017}
Aaron~J Cohen, Michael Brauer, Richard Burnett, H~Ross Anderson, Joseph
  Frostad, Kara Estep, Kalpana Balakrishnan, Bert Brunekreef, Lalit Dandona,
  Rakhi Dandona, Valery Feigin, Greg Freedman, Bryan Hubbell, Amelia Jobling,
  Haidong Kan, Luke Knibbs, Yang Liu, Martin Randall, and Mohammad~H
  Forouzanfar.
\newblock Estimates and 25-year trends of the global burden of disease
  attributable to ambient air pollution: an analysis of data from the {Global}
  {Burden} of {Diseases} {Study} 2015.
\newblock \emph{The Lancet}, 389\penalty0 (10082):\penalty0 1907--1918, May
  2017.
\newblock ISSN 0140-6736.
\newblock \doi{10.1016/S0140-6736(17)30505-6}.
\newblock URL
  \url{https://www.sciencedirect.com/science/article/pii/S0140673617305056}.

\bibitem[Lelieveld et~al.(2015)Lelieveld, Evans, Fnais, Giannadaki, and
  Pozzer]{lelieveld_contribution_2015}
J.~Lelieveld, J.~S. Evans, M.~Fnais, D.~Giannadaki, and A.~Pozzer.
\newblock The contribution of outdoor air pollution sources to premature
  mortality on a global scale.
\newblock \emph{Nature}, 525\penalty0 (7569):\penalty0 367--371, September
  2015.
\newblock ISSN 1476-4687.
\newblock \doi{10.1038/nature15371}.
\newblock URL \url{https://www.nature.com/articles/nature15371}.
\newblock Number: 7569 Publisher: Nature Publishing Group.

\bibitem[Carminati et~al.(2017)Carminati, Ferrari, and
  Sampietro]{carminati_emerging_2017}
Marco Carminati, Giorgio Ferrari, and Marco Sampietro.
\newblock Emerging miniaturized technologies for airborne particulate matter
  pervasive monitoring.
\newblock \emph{Measurement}, 101:\penalty0 250--256, April 2017.
\newblock ISSN 0263-2241.
\newblock \doi{10.1016/j.measurement.2015.12.028}.
\newblock URL
  \url{http://www.sciencedirect.com/science/article/pii/S0263224115006880}.

\bibitem[Liu et~al.(2020)Liu, Jayaratne, Thai, Kuhn, Zing, Christensen, Lamont,
  Dunbabin, Zhu, Gao, Wainwright, Neale, Kan, Kirkwood, and
  Morawska]{liu_low-cost_2020}
Xiaoting Liu, Rohan Jayaratne, Phong Thai, Tara Kuhn, Isak Zing, Bryce
  Christensen, Riki Lamont, Matthew Dunbabin, Sicong Zhu, Jian Gao, David
  Wainwright, Donald Neale, Ruby Kan, John Kirkwood, and Lidia Morawska.
\newblock Low-cost sensors as an alternative for long-term air quality
  monitoring.
\newblock \emph{Environmental Research}, 185:\penalty0 109438, June 2020.
\newblock ISSN 0013-9351.
\newblock \doi{10.1016/j.envres.2020.109438}.
\newblock URL
  \url{http://www.sciencedirect.com/science/article/pii/S0013935120303315}.

\bibitem[Smith et~al.(2019)Smith, Ssematimba, Alvarez, and
  Bainomugisha]{smith_machine_2019}
Michael~T. Smith, Joel Ssematimba, Mauricio~A. Alvarez, and Engineer
  Bainomugisha.
\newblock Machine {Learning} for a {Low}-cost {Air} {Pollution} {Network}.
\newblock \emph{NeurIPS 2019 Workshop on Machine Learning for the Developing
  World}, December 2019.
\newblock URL \url{http://arxiv.org/abs/1911.12868}.
\newblock arXiv: 1911.12868.

\bibitem[Liu et~al.(2016)Liu, Xi, and Ngai]{liu_data_2016}
Xiuming Liu, Teng Xi, and Edith Ngai.
\newblock Data {Modelling} with {Gaussian} {Process} in {Sensor} {Networks} for
  {Urban} {Environmental} {Monitoring}.
\newblock In \emph{2016 {IEEE} 24th {International} {Symposium} on {Modeling},
  {Analysis} and {Simulation} of {Computer} and {Telecommunication} {Systems}
  ({MASCOTS})}, pages 457--462, September 2016.
\newblock \doi{10.1109/MASCOTS.2016.45}.
\newblock ISSN: 2375-0227.

\bibitem[Liu et~al.(2018)Liu, Yang, Huang, Wang, and Yoo]{liu_modeling_2018}
Hongbin Liu, Chong Yang, Mingzhi Huang, Dongsheng Wang, and ChangKyoo Yoo.
\newblock Modeling of subway indoor air quality using {Gaussian} process
  regression.
\newblock \emph{Journal of Hazardous Materials}, 359:\penalty0 266--273,
  October 2018.
\newblock ISSN 0304-3894.
\newblock \doi{10.1016/j.jhazmat.2018.07.034}.
\newblock URL
  \url{http://www.sciencedirect.com/science/article/pii/S0304389418305648}.

\bibitem[Fei et~al.(2020)Fei, Wu, Luo, and Luo]{fei_accurate_2020}
Haolin Fei, Xiaofeng Wu, Chunbo Luo, and Yang Luo.
\newblock Accurate {Air} {Quality} {Prediction}: {A} {Physical}-{Temporal}
  {Collection} {Model}.
\newblock In \emph{{ICLR} {Workshop} on {AI} for {Earth} {Science}}, 2020.
\newblock URL
  \url{https://www.researchgate.net/publication/341445468_ACCURATE_AIR_QUALITY_PREDICTION_A_PHYSICAL-_TEMPORAL_COLLECTION_MODEL_CONFERENCE_SUBMISSIONS}.

\bibitem[Wen et~al.(2019)Wen, Liu, Yao, Peng, Li, Hu, and Chi]{wen_novel_2019}
Congcong Wen, Shufu Liu, Xiaojing Yao, Ling Peng, Xiang Li, Yuan Hu, and Tianhe
  Chi.
\newblock A novel spatiotemporal convolutional long short-term neural network
  for air pollution prediction.
\newblock \emph{Science of The Total Environment}, 654:\penalty0 1091--1099,
  March 2019.
\newblock ISSN 0048-9697.
\newblock \doi{10.1016/j.scitotenv.2018.11.086}.
\newblock URL
  \url{http://www.sciencedirect.com/science/article/pii/S0048969718344413}.

\bibitem[Marchant and Ramos(2012)]{marchant_bayesian_2012}
R.~Marchant and F.~Ramos.
\newblock Bayesian optimisation for {Intelligent} {Environmental} {Monitoring}.
\newblock In \emph{2012 {IEEE}/{RSJ} {International} {Conference} on
  {Intelligent} {Robots} and {Systems}}, pages 2242--2249, October 2012.
\newblock \doi{10.1109/IROS.2012.6385653}.

\bibitem[Morere et~al.(2017)Morere, Marchant, and
  Ramos]{morere_sequential_2017}
Philippe Morere, Roman Marchant, and Fabio Ramos.
\newblock Sequential {Bayesian} optimization as a {POMDP} for environment
  monitoring with {UAVs}.
\newblock In \emph{2017 {IEEE} {International} {Conference} on {Robotics} and
  {Automation} ({ICRA})}, pages 6381--6388, May 2017.
\newblock \doi{10.1109/ICRA.2017.7989754}.

\bibitem[Singh et~al.(2010)Singh, Ramos, Whyte, and
  Kaiser]{singh_modeling_2010}
A.~Singh, F.~Ramos, H.~D. Whyte, and W.~J. Kaiser.
\newblock Modeling and decision making in spatio-temporal processes for
  environmental surveillance.
\newblock In \emph{2010 {IEEE} {International} {Conference} on {Robotics} and
  {Automation}}, pages 5490--5497, May 2010.
\newblock \doi{10.1109/ROBOT.2010.5509934}.

\bibitem[Samson(1988)]{watson_atmospheric_1988}
Perry~J Samson.
\newblock Atmospheric {Transport} and {Dispersion} of {Air} {Pollutants}
  {Associated} with {Vehicular} {Emissions}.
\newblock In Ann~Y. Watson, Richard~R. Bates, and Donald Kennedy, editors,
  \emph{Air {Pollution}, the {Automobile}, and {Public} {Health}}. National
  Academies Press (US), 1988.
\newblock URL \url{https://www.ncbi.nlm.nih.gov/books/NBK218142/}.
\newblock Publication Title: Air Pollution, the Automobile, and Public Health.

\bibitem[Reggente and Lilienthal(2009)]{reggente_using_2009}
Matteo Reggente and Achim~J. Lilienthal.
\newblock Using local wind information for gas distribution mapping in outdoor
  environments with a mobile robot.
\newblock In \emph{2009 {IEEE} {SENSORS}}, pages 1715--1720, October 2009.
\newblock \doi{10.1109/ICSENS.2009.5398498}.
\newblock ISSN: 1930-0395.

\bibitem[Asenov et~al.(2019)Asenov, Rutkauskas, Reid, Subr, and
  Ramamoorthy]{asenov_active_2019}
Martin Asenov, Marius Rutkauskas, Derryck Reid, Kartic Subr, and Subramanian
  Ramamoorthy.
\newblock Active {Localization} of {Gas} {Leaks} {Using} {Fluid} {Simulation}.
\newblock \emph{IEEE Robotics and Automation Letters}, 4\penalty0 (2):\penalty0
  1776--1783, April 2019.
\newblock ISSN 2377-3766.
\newblock \doi{10.1109/LRA.2019.2895820}.
\newblock Conference Name: IEEE Robotics and Automation Letters.

\bibitem[Luo et~al.(2016)Luo, Li, Li, Wang, Ma, Zhang, Fang, Wu, Cao, and
  Xu]{luo_acute_2016}
Kai Luo, Runkui Li, Wenjing Li, Zongshuang Wang, Xinming Ma, Ruiming Zhang, Xin
  Fang, Zhenglai Wu, Yang Cao, and Qun Xu.
\newblock Acute {Effects} of {Nitrogen} {Dioxide} on {Cardiovascular}
  {Mortality} in {Beijing}: {An} {Exploration} of {Spatial} {Heterogeneity} and
  the {District}-specific {Predictors}.
\newblock \emph{Scientific Reports}, 6\penalty0 (1):\penalty0 38328, December
  2016.
\newblock ISSN 2045-2322.
\newblock \doi{10.1038/srep38328}.
\newblock URL \url{https://www.nature.com/articles/srep38328}.
\newblock Number: 1 Publisher: Nature Publishing Group.

\bibitem[Shahriari et~al.(2016)Shahriari, Swersky, Wang, Adams, and
  de~Freitas]{shahriari_taking_2016}
Bobak Shahriari, Kevin Swersky, Ziyu Wang, Ryan~P. Adams, and Nando de~Freitas.
\newblock Taking the {Human} {Out} of the {Loop}: {A} {Review} of {Bayesian}
  {Optimization}.
\newblock \emph{Proceedings of the IEEE}, 104\penalty0 (1):\penalty0 148--175,
  January 2016.
\newblock ISSN 1558-2256.
\newblock \doi{10.1109/JPROC.2015.2494218}.
\newblock Conference Name: Proceedings of the IEEE.

\bibitem[Srinivas et~al.(2012)Srinivas, Krause, Kakade, and
  Seeger]{srinivas_gaussian_2012}
Niranjan Srinivas, Andreas Krause, Sham~M. Kakade, and Matthias Seeger.
\newblock Gaussian {Process} {Optimization} in the {Bandit} {Setting}: {No}
  {Regret} and {Experimental} {Design}.
\newblock \emph{IEEE Transactions on Information Theory}, 58\penalty0
  (5):\penalty0 3250--3265, May 2012.
\newblock ISSN 0018-9448, 1557-9654.
\newblock \doi{10.1109/TIT.2011.2182033}.
\newblock URL \url{http://arxiv.org/abs/0912.3995}.
\newblock arXiv: 0912.3995.

\bibitem[Hernández-Lobato et~al.(2014)Hernández-Lobato, Hoffman, and
  Ghahramani]{hernandez-lobato_predictive_2014}
José~Miguel Hernández-Lobato, Matthew~W Hoffman, and Zoubin Ghahramani.
\newblock Predictive {Entropy} {Search} for {Efficient} {Global} {Optimization}
  of {Black}-box {Functions}.
\newblock \emph{Advances in Neural Information Processing Systems 27}, pages
  918--926, 2014.

\bibitem[Rasmussen and Williams(2006)]{rasmussen_gaussian_2006}
Carl~Edward Rasmussen and Christopher K.~I. Williams.
\newblock \emph{Gaussian {Processes} for {Machine} {Learning}}.
\newblock MIT Press, 2006.

\bibitem[Cats and Holtslag(1980)]{cats_prediction_1980}
GJ~Cats and AAM Holtslag.
\newblock Prediction of air pollution frequency distribution—{Part} {I}.
  {The} lognormal model.
\newblock \emph{Atmospheric Environment (1967)}, 14:\penalty0 255--258, 1980.
\newblock URL
  \url{https://www.sciencedirect.com/science/article/pii/0004698180902851}.

\bibitem[Schwarz(1978)]{schwarz_estimating_1978}
Gideon Schwarz.
\newblock Estimating the dimension of a model.
\newblock \emph{The annals of statistics}, 6\penalty0 (2):\penalty0 461--464,
  1978.

\end{thebibliography}
\bibliographystyle{unsrtnat}}

\newpage


\appendix
\section{Supplementary Material}
Supplementary material for \emph{Optimising Placement of Pollution Sensors in Windy Environments} by Sigrid Passano Hellan, Christopher G. Lucas and Nigel H. Goddard presented at the AI for Earth Sciences Workshop at NeurIPS 2020.

\subsection{Data} \label{app-data}

The data were collected by the TROPOspheric Measuring Instrument (TROPOMI) aboard the Sentinel 5P satellite,\footnote{Copernicus, \url{https://www.copernicus.eu/en}}
and were extracted and shared with us by 
Dr Douglas Finch at the University of Edinburgh.
In the original data source, the \NOt data used is given in the column labelled 'level 2 nitrogen dioxide tropospheric'.
Images covering mostly oceans were excluded. 

\subsection{BIC scores} \label{app-likelihoods}

The Bayesian information criterion (BIC) scores of the tuning sets referred to in Section 3 are given in \cref{tab:oracle-likelihoods}. The values are given as differences because the obtained BIC scores varied a lot between problems, e.g. between 306 and -1014 for the Sum kernel on the Strong subset. As can be seen, the relative differences vary a lot less.

\begin{table}[ht]
\centering
\caption{Differences in Bayesian information criterion (BIC) scores obtained on tuning sets. Mean (standard deviation of mean). The wind-informed kernels were better able to explain the data for the Strong and Median subsets. There was little difference for the Weak subset, where the images were very noisy.}
\label{tab:oracle-likelihoods}
\begin{tabular}{@{}llll@{}}
\toprule
            & Sum - Product & Sum - RBF    & Product - RBF \\ \midrule
Strong tune & 13.43 (3.69)  & 19.92 (3.44) & 6.49 (2.75)  \\
Median tune & 2.69 (1.95)   & 10.02 (4.30) & 7.33 (3.73)  \\
Weak tune & 0.35 (0.58)   & -1.12 (1.72) & -1.47 (1.34)  \\ \bottomrule
\end{tabular}
\end{table}

\subsection{Implementation notes}

A small amount of noise was added to sampled values in the BO loop for numerical stability. 
At each iteration in the BO loop 100 hyperparameter seeds were sampled from the priors and used as starting points for the GP tuning. The setting which maximised the posterior log marginal likelihood was then used. 
When estimating the hyperparameter priors, Bessel's correction was used for the variance estimate.

\subsection{Hyperparameter prior sensitivity}

\begin{figure}[ht]
\centering
\begin{subfigure}[b]{.47\textwidth}
  \centering
  \includegraphics[width=.95\linewidth]{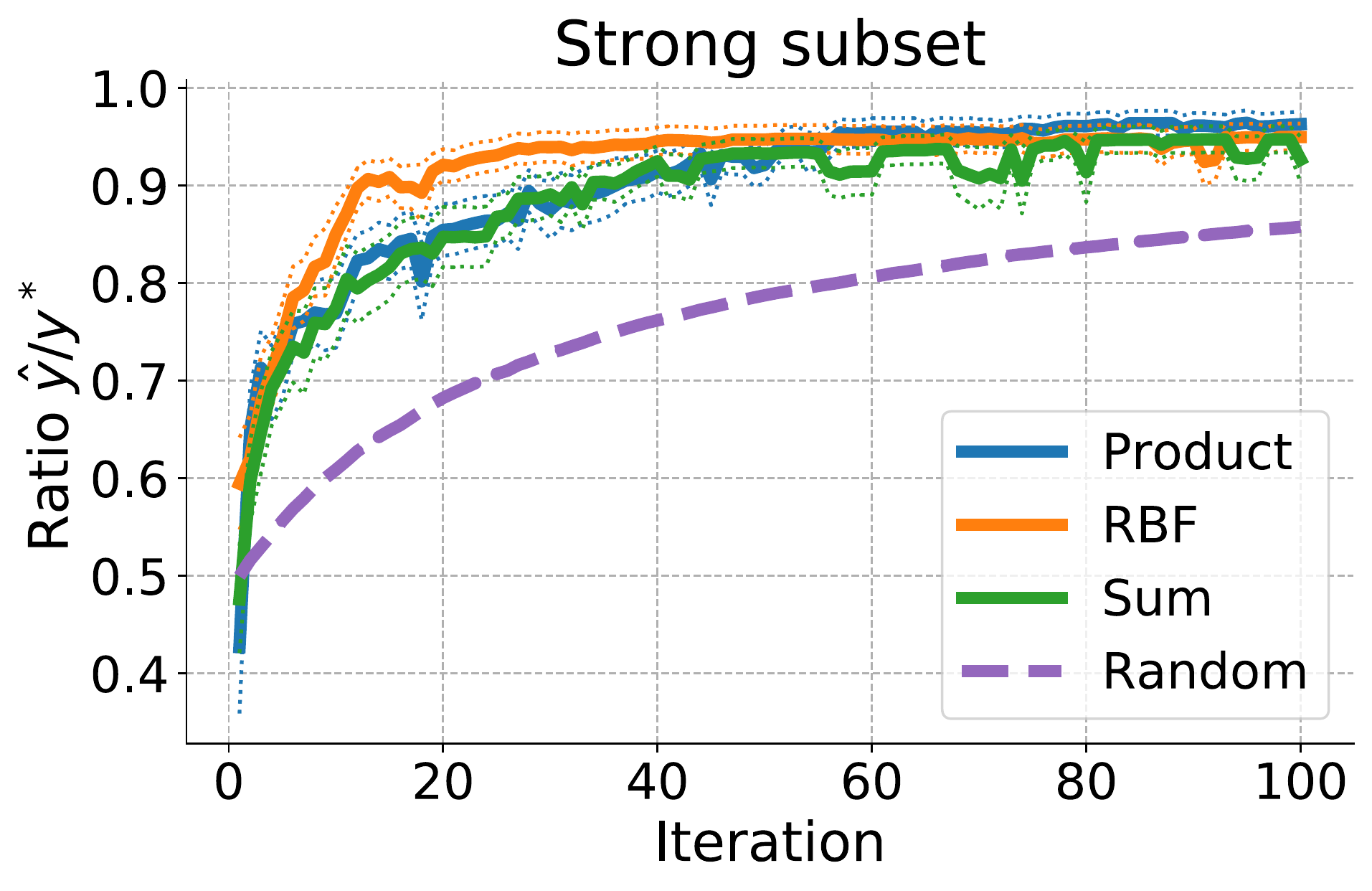}
  \label{fig:results-strong-true-appendix}
\end{subfigure}%
\begin{subfigure}[b]{.47\textwidth}
  \centering
  \includegraphics[width=.95\linewidth]{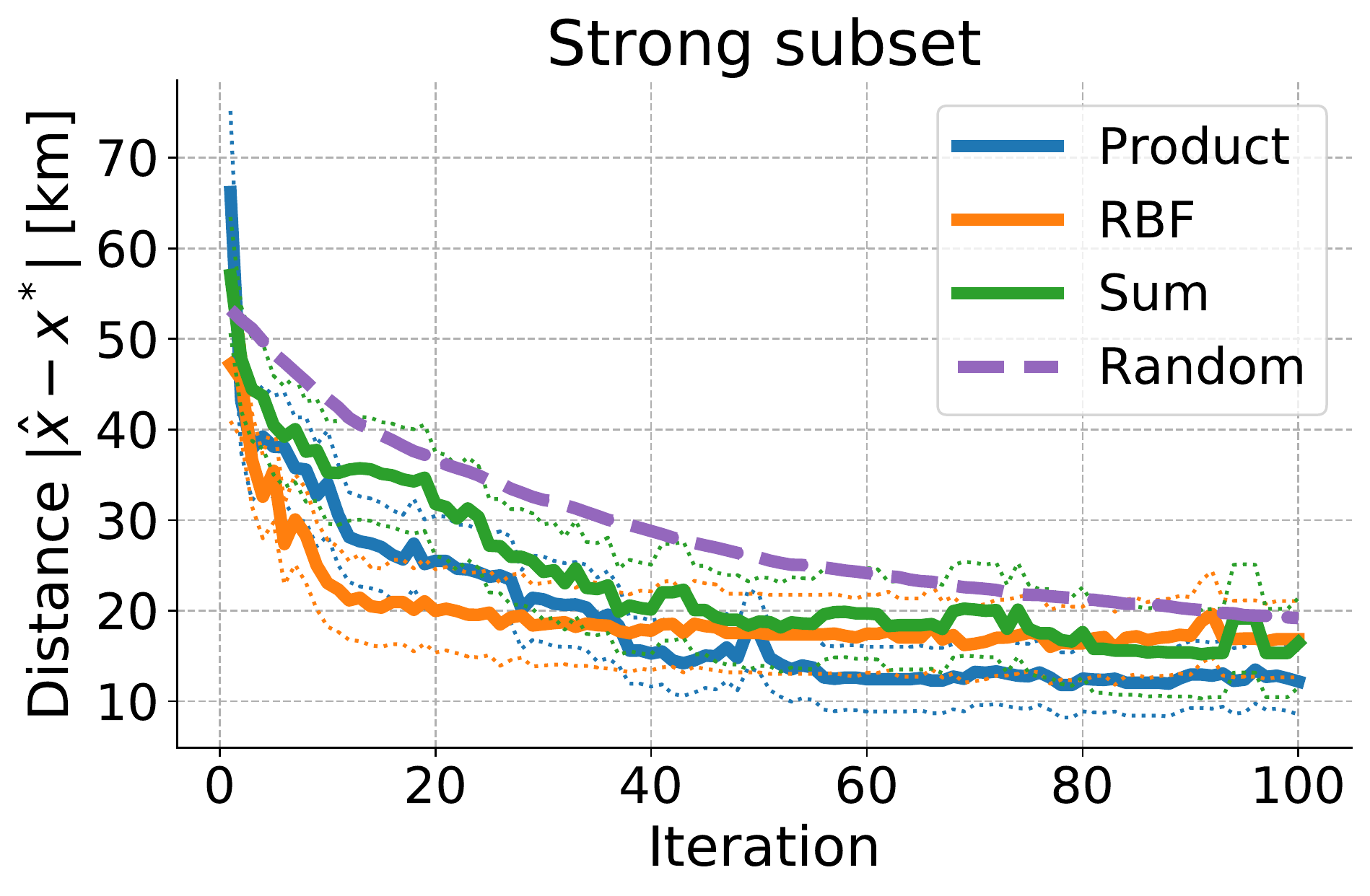}
  \label{fig:results-strong-dist-appendix}
\end{subfigure}%
\caption{Comparison of kernel performance on Strong subsets, when biased estimates of variances of hyperparameter priors are  used, i.e. not using Bessel's correction. Compare to Fig. 3 in body of paper. With these settings, the Product kernel performs best. Comparing the figures, the Product kernel has improved but the RBF worsened. $\hat{x}$ is the estimated maximiser and $x^*$ the true maximiser. $\hat{y}$ and $y^*$ are the true concentration values at $\hat{x}$ and $x^*$, respectively. }
\label{fig:results-strong-appendix}
\end{figure}

\cref{fig:results-strong-appendix} shows the results on the Strong subset when a slightly different prior was used for the hyperparameters, as referred to in Section 4 of the main paper. 

\end{document}